\documentclass[10pt,twocolumn,letterpaper]{article}
\usepackage{iccv}
\usepackage{times}
\usepackage{epsfig}
\usepackage{graphicx}
\usepackage{amsmath}
\usepackage{amssymb}
\usepackage{siunitx}

\usepackage[pagebackref=true,breaklinks=true,letterpaper=true,colorlinks,bookmarks=false]{hyperref}

\iccvfinalcopy 

\def\iccvPaperID{2221} 

\ificcvfinal\fi

\usepackage{times}
\usepackage{epsfig}
\usepackage{graphicx}
\usepackage{amsmath}
\usepackage{amssymb}
\usepackage{mathtools}
\usepackage[utf8]{inputenc} 
\usepackage[T1]{fontenc}    
\usepackage{hyperref}       
\usepackage{url}            
\usepackage{booktabs}       
\usepackage{paralist}       
\usepackage{amsfonts}       
\usepackage{nicefrac}       
\usepackage{microtype}      
\usepackage{graphicx}
\usepackage[ruled,vlined]{algorithm2e}

\usepackage{floatrow}
\newfloatcommand{capbtabbox}{table}[][\FBwidth]

\usepackage{blindtext}
\def\HH{\mathbb{H}}
\def\E{I\!\!E}

\begin{document}

\title{BRUL\`E: Barycenter-Regularized Unsupervised Landmark Extraction}

\author{\\
Iaroslav Bespalov\thanks{Contributed equally.},~~~Nazar Buzun\footnotemark[1],~~~Dmitry V. Dylov\thanks{Corresponding author.}\\
Skolkovo Institute of Science and Technology\\
Bolshoy blvd., 30/1, Moscow, Russia 121205\\
{\{iaroslav.bespalov, n.buzun, d.dylov\}@skoltech.ru}
}


\maketitle
\ificcvfinal\thispagestyle{empty}\fi

\begin{abstract}
Unsupervised retrieval of image features is vital for many computer vision tasks where the annotation is missing or scarce. 
In this work, we propose a new unsupervised approach to detect the landmarks in images, validating it on the popular task of human face key-points extraction. 
The method is based on the idea of auto-encoding the wanted landmarks in the latent space while discarding the non-essential information (and effectively preserving the interpretability). 
The interpretable latent space representation (the bottleneck containing nothing but the wanted key-points) is achieved by a new two-step regularization approach. 
The first regularization step evaluates transport distance from a given set of landmarks to some average value (the barycenter by Wasserstein distance). 
The second regularization step controls deviations from the barycenter by applying random geometric deformations synchronously to the initial image and to the encoded landmarks. 
We demonstrate the effectiveness of the approach both in unsupervised and semi-supervised training scenarios using 300-W, CelebA, and MAFL datasets. 
The proposed regularization paradigm is shown to prevent overfitting, and the detection quality is shown to improve beyond the state-of-the-art face models.
\end{abstract}

\section{Introduction}

Our study of the unsupervised landmark detection began with the question of whether it is possible to store the image landmarks within the bottleneck of an auto-encoder.
How can we influence its bottleneck to contain \emph{only} the information about the image landmarks and nothing else? What kind of regularization would be required for that?

Using auto-encoders to extract landmarks is aligned with the \textit{ultimate vision} of unsupervised segmentation, because it is in the bottleneck where the features of the unlabeled contours could be distilled~\cite{8,7}. Image segmentation (with varying extent of supervision) has been one of the most popular tasks in the field of deep learning over the last five years~\cite{1,2,3,6}. Today, the state-of-the-art algorithms show impressive results but, oftentimes, require large volumes of annotated data~\cite{2,3}. Alleviating the annotation burden has been a task of pressing demand and is the other motivation for us to search for the unsupervised solutions. We begin the effort with the detection of key-points in the human faces, a simpler task than generic segmentation and the main focus of this paper.

The questions mentioned above, along with the fact that the landmarks of the same class typically look similar\footnote{\textit{E.g.}, key-points extracted from different faces resemble each other.}, have led us to the idea of comparing the post-encoder features with some `average' landmarks.
Naturally, such an `average' pattern could be computed by the optimal transport (OT) distance~\cite{main_bc_ref}, also known as the \emph{Wasserstein barycenter}. The OT distance evaluates the size of geometric deformation needed to transform one set of landmarks to another, effectively making the `average' pattern look natural.
Wasserstein barycenters rose in popularity in recent years as they preserve common topological properties of geometric objects for various computer vision tasks~\cite{bc_coords_proj,bc_gar,nikoshvez}.
Herein, we extend their applicability to the landmark extraction problem. 

Interestingly, the predicted landmarks may contain some information about the face position and its contours, but there still may be no direct correspondence between the landmark value $(\in [0, 1]^2)$ and the coordinate in the original face image~\cite{WingLoss}. Synchronization of coordinates by \textit{geometric transforms}, proposed in papers \cite{8, DVE}, has motivated us to consider a second regularizer. 
Specifically, we propose to synchronously deform the image and its predicted landmarks, predict landmarks of the deformed image and, finally, compare them to the transformed landmarks. Such coordinate synchronization would control deviations from the barycenter, which has never been attempted before. 

\begin{figure*}[hbt!]
\begin{center}
\includegraphics[width=0.85\textwidth]{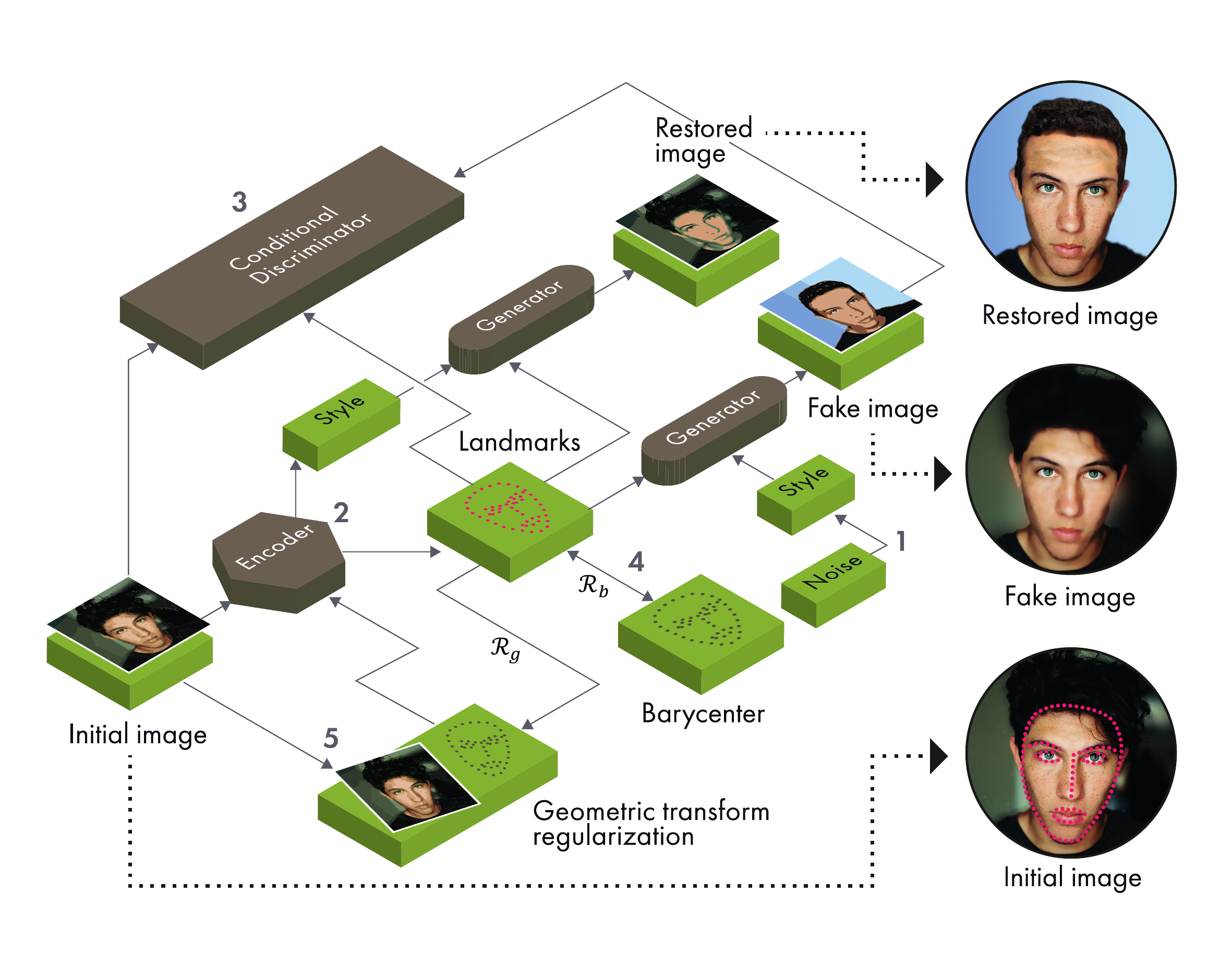}
\caption{BRUL\`E: Barycenter-Regularized Unsupervised Landmark Extraction. (1) The first restoration generates style from noise and forwards it to GAN. (2) The second restoration predicts style and landmarks from the initial image and applies the same adversarial generator. (3) The initial and the restored images are compared by conditional discriminator and the $L_1$ norm. Wasserstein-2 distance forces the encoded landmarks to be close to barycenter. (4) The image and its predicted landmarks are synchronously deformed via affine and elastic geometric transforms. (5) Finally, one predicts the landmarks of the transformed image and compares them to the transformed landmarks.}
 \label{fig:architecture}
\end{center}
\end{figure*}

The proposed architecture for the barycenter regularization is shown in Fig.~\ref{fig:architecture} and is discussed in detail in Section \ref{loss_func_sec}. 
For decoding, we propose to use a generative adversarial network (GAN), constructed as a powerful combination of modified \texttt{stylegan2}~\cite{Stylegan2} and \texttt{MUNIT}~\cite{MUNIT} architectures (details are given in Section~\ref{method_sec}).

The contribution of this paper is in the following:
\begin{compactitem}
\item The first method that predicts \textbf{interpretable} landmarks in \textbf{unsupervised} way. 
\item Unlike pre-trained models which require large datasets for \textit{pre}-training their auto-encoders, our method needs \textbf{just a dozen of images} to compute barycenter.
\item In a \textit{semi}-supervised scenario, our method \textbf{outperforms state-of-the-art} models.
\item Two types of regularization (\textit{barycenter} and \textit{geometric transforms}) are shown to suffice for auto-encoder to produce the image landmarks \textit{in the bottleneck}.
\item New type of cyclic/conditional GAN architecture \cite{cyclegan} that performs training with only one domain data and decomposes images into landmarks and style.
\end{compactitem}

%
\section{Related Work}
Traditionally, the algorithms of the unsupervised segmentation extract latent representations via deep autoencoders~\cite{Chapelle-semisup-review,8,7,6,shu2018deforming}. 
These methods attempt to form clusters of the latent vectors which correspond to correlated parts of the initial image.   
To guarantee direct correspondence between image and segmentation coordinates, the authors in~\cite{8} suggested the idea of regularization using geometry transformations, expressed as a condition $L(gI) = gL(I)$, where $I$ is the image, $g$ is some random deformation, and $L$ is a segmentation mapping. \emph{GANs}, such as SEIGAN~\cite{9}, were also proposed for unsupervised segmentation, relying on the latent space representation, segment painting, and object embedding into another background (with the constraint that the image must remain realistic).

There are many landmark detection approaches, especially for faces. Initially, they were based on active appearance and entailed various statistical approaches, pattern matching, preprocessing, filtering, and deformations~\cite{AAM, AFLCLM, JAIN1998-DeformReview,WIDROW1973-rubber}. Rapid progress of deep learning then instigated a series of supervised methods: cascade of CNNs~\cite{DCNCFPD}, multi-task learning (pose, smile, the presence of glasses, sex of person)~\cite{TCDCN}, and recurrent attentive-refinements via Long Short-Term Memory networks (LSTMs)~\cite{RAR}. Special loss functions (\textit{e.g.}, wing loss~\cite{WingLoss}) were shown to further improve the accuracy of CNN-based facial landmark localisation.

\emph{Unsupervised pre-training} has seen major interest in the community with the advent of data-hungry deep networks~\cite{DVE}. A classic approach for such a task is to use the geometric transformations for regularization, comprising different variations of embeddings~\cite{Dense3D, SPARSE, DVE}. 
However, none of these works reports extraction of the landmarks in a truly unsupervised way~\cite{Qi-Unsup-Semisup-Review}. Their unsupervised nature is only as good as a generic \emph{pre-training} could be, still requiring \textbf{large datasets to pre-train encoders} and being prone to encoding some \textbf{redundant information within the bottleneck} (and not \emph{just} the landmark/segmentation data).

Another class of unsupervised pre-training methods~\cite{kp_next_im_gen, FabNet,Jakab_ConditionalGeneration} use the condition $I_2 = G(I_1, L(I_2))$ ($I_1$ and $I_2$ are two images) and the additional condition of sparsity on the heatmap corresponding to $L(I_2)$. If $I_1$ and $I_2$ have different landmarks but have the same style (\textit{e.g.}, sequential frames from the video), the network $G$ can generate $I_2$ from $I_1$ and the landmarks $L(I_2)$ of the second image~$I_2$. 
Method~\cite{UDIT} is similar, but it has an additional discriminator network to compare predicted landmarks to the landmarks from some unaligned dataset by distribution. Besides the added complexity, the use of unaligned dataset could be viewed as an \textit{intermediate scenario} between the unsupervised and the supervised training~\cite{Chapelle-semisup-review}. To the contrary, we show that instead of relying on the unaligned models, one may simply combine the proposed barycenter regularizer with the geometric transforms, yielding the true unsupervised functionality.  

Lastly, in our approach, we have chosen Wasserstein distance because it can establish \textit{pairwise correspondence} between the predicted landmarks and the key-points of the barycenter. It makes the regularizer more flexible, enabling the capability of comparing the landmark sets of different sizes.  We refer to works~\cite{arjovsky2017wasserstein,main_bc_ref, bc_gar, Feydy-Sinkhorn} that describe useful theoretical properties of Wasserstein barycenters.
\section{Method: BRUL\`E}
\label{loss_func_sec}
\noindent We begin by describing 5 principal ingredients of BRUL\`E architecture shown in Fig.~\ref{fig:architecture}, with each of them having its own \emph{physical meaning} and the corresponding loss term. 
%
\paragraph{Barycenter Regularizer $\mathcal{R}_{\flat}$.} Let $X^{\flat} \in [0, 1]^{2N}$ be the coordinates of the barycenter and $X^{l} \in [0, 1]^{2N}$ be the coordinates of the predicted landmarks, where $N$ is the total number of the corresponding 2D points.   
We compare $X^{\flat}$ with $X^{l}$  by means of transport (Wasserstein-2) distance \cite{villani} denoted as $W_2^2$.  The transport path from $X^{l}$ to the barycenter $X^{\flat}$ entails two principal transformations: the linear (affine) and the nonlinear (warping). Hence, the transport mapping is expressed as sequential translation $T$, two other affine transformations $A$ (rotation and scaling), and the nonlinear elastic deformation\footnote{We find expressing the translation term separately from the other affine transformations to be useful due to its lesser impact on the regularization.}:
\begin{align*}
&\mathcal{R}_{\flat}(X^{l}) = c_{1}^{\flat} W^2_2\left(X^{l}, T[X^{l}]\right)  \\
& + c_{2}^{\flat} W_2^2\left(T[X^{l}],
A\,T[X^{l}]\right) + c_{3}^{\flat} W_2^2\left(A\,T[X^{l}], X^{\flat}\right),
\end{align*}

\noindent where $c_{1}^{\flat}$, $c_{2}^{\flat}$, and $c_{3}^{\flat}$ are the coefficients to vary the strength of the regularization by each term. Naturally, simpler deformations are preferred, yielding $c_{1}^{\flat} < c_{2}^{\flat} < c_{3}^{\flat}$.

The translation $T[X^{l}]$ is determined by the center of mass difference between $X^{l}$ and $X^{\flat}$. Conventionally, one could express the affine matrix $A$ via the covariance matrix $\Sigma$; however, it would have given a solution up to an orthogonal matrix, because $\Sigma = A\,A^T$ and one might inject the orthogonal transformation between $A$ and $A^T$.
It is possible to resolve this irregularity by establishing a \textit{pairwise correspondence} between the source and the destination of the linear transport mapping. Particularly, if $P$ is a probability matrix of the complete transport path, such that $P_{ij}$ is the probability that $i$-th point from $X^{l}$ moves to $j$-th point of $X^{\flat}$, one can find the matrix of affine operator $A$ by solving the following optimization problem: 
\begin{equation*}
    \min_{A}\Big \{ \sum_{ij = 1}^N  P_{ij} \big\| ATX^{l}_i - X^{\flat}_j \big\|^2 \Big \},
\end{equation*}
yielding the solution:
\begin{equation}
    A = \left( (TX^{l})^T \text{diag}(P\vec{1}) TX^{l}\right)^{-1} (TX^{l})^T P X^{\flat} .
\end{equation}

\paragraph{Geometric Regularizer $\mathcal{R}_{g}$.} To guarantee correspondence of coordinates of the landmarks to those in the image, we add a proper geometric regularization. For that, in addition to the landmark coordinates  $X^{l}$, we engage Gaussian \textit{heatmaps} $L = L(X^{l})$ which are computed as follows. Let $E$ be the edges set of a $k$-nearest neighbors graph ($k=2$), built on $X^{l}$. For an edge $e$ and integer coordinates $(i,j)$ in the heatmap, denote the distance from $(i,j)$ to the nearest point in $e$ by $d(e, i, j)$. Then, up to the normalisation term,  
\begin{equation}
\label{hm_eq}
L_{i j} = \sum_{e \in E} \exp\left\{ - \frac{d^2(e, i, j)}{2 \sigma^2} \right\}.
\end{equation}
Visually, such heatmaps $L$ look like a blurred skeleton.  The geometric regularization assures that the same affine and elastic transformations are applied both to the original image $I$ and to the predicted landmarks $L$. 

After applying the deformations $g$, we use the encoder to predict a new set of landmarks $L(gI)$ of the transformed image $gI$. 
So, the following loss component will minimize cross-entropy $\mathbb{H}$ between $L(gI)$ and $gL$, along with the $l_1$ distance between their coordinates:
\[
    \mathcal{R}_g(I, L) =  \mathbb{H} \left[ L(gI) \, | \, gL \right] + c_1^g \, \| X^{l}(gI) - gX^{l} \|_1 ,
\]
where $c_1^g$ controls the influence of $l_1$ norm.

\paragraph{Reconstruction Loss.} Let us denote the generator (decoder) model by $G$. It maps a pair (\textit{landmarks}, \textit{style}) into an image, where the \textit{style} may be either encoded from the image $S(I)$ or generated from the noise $S_o(z)$.  Then, this loss term compares the original image to the reconstructed one, for a given heatmap of landmarks and the encoded style:
\[
    \mathcal{L}_{rec}^{I}( L, G, S) = \|  G(L, S(I)) - I  \|_1 .
\]
Contrary to $ \mathcal{R}_g$, this loss term enforces the encoder to extract as much information as possible, preventing collapse of the nearest points in a given landmark set. Similarly, for the landmarks reconstruction obtained from a \textit{fake} image:
\[
 \mathcal{L}_{rec}^{L}(L) = \HH [ L(I_F) \,| \, L ] ,
\]
where $I_F = G(L, S_o(z))$.
\paragraph{Adversarial Loss.} We further increase correlation between the image and the landmarks by chaining with the GAN losses \cite{Stylegan2} for discriminator and generator:
\[
\mathcal{L}_d(D) = -\E\log\big(1 + e^{-D(I, L)}\big)-\E \log \big(1 + e^{D(I_F,L)}\big),
\]
\[
\mathcal{L}_g(I_F) = \E\log\left(1 + e^{-D(I_F, L)}\right).
\]
This loss component acts similarly to $\mathcal{L}_{rec}^{I}$, but is more flexible because it does not depend on the style encoding $S(I)$.  

\paragraph{Style Consistency Loss.} We make style consistent with the style generated from noise and make it invariant to the geometric transformations:
\[
\mathcal{L}_{\text{style}}(S) =  \| S - S_o(z) \|_1 +  \| S - S(gI_F) \|_1 .
\]
\section{Training Algorithms}
\label{method_sec}

The training process of the unsupervised landmark detector, shown in Fig.~\ref{fig:architecture}, consists of two main steps: conditional GAN training and the actual landmark detection optimization. These two steps are repeated every training iteration and presented in Algorithms~\ref{gan_train} and \ref{encoder_train} correspondingly. When the encoder network predicts the heatmap of landmarks $L$ and the style $S$ of the batch of images~$I$, one applies the optimization routine to the generator ($G$) and the discriminator ($D$) networks. Recall that $S_o$ denotes the network that maps the Gaussian noise vector $z$ to a random style, and $S$ is the network that maps images (either fake or real) to their style.

\begin{algorithm}[t]
\SetAlgoLined
  \textbf{Input}: image $I$, landmarks $L$\;
 sample noise: $z_k \sim \mathcal{N}(0, 1)$,  $\quad k \in \{1,\ldots,K\}$\;
 initialize $g$: random geometric transformation\;
 generate fake image: $I_F = G(L, S_o(z))$\;
 $\max_D$ discriminator loss: $\mathcal{L}_d(D)$\;
 \If{~iteration$\mod 4 == 0$} {
    $\min_D$ discriminator penalty : $\lambda\E \|\nabla D(I,L)\|^{2} $\;
 }
 
 $\min_{G}$ generator loss: $\mathcal{L}_g(I_F)$\;
 restore image: $I_R = G(L, S(I))$\;
 style of fake: $S_F = S(I_F)$\;
    $\min_{G, S} \left \{ 
      c_1  \mathcal{L}_{rec}^{I}(I_R, I) + c_2 \mathcal{L}_{style}(S_F)
    \right \}$
 \caption{Single iteration of conditional GAN optimization}
 \label{gan_train}
\end{algorithm}

The training routine of Algorithm~\ref{gan_train} starts similarly to MUNIT~\cite{MUNIT}, where one encodes the style and the landmarks from an input image. Then, we deviate by separately restoring the image and generating the fake. 
It makes GAN training more stable and allows to decompose the image into content and style, with the role of content being played by the landmarks.
The loss function for the discriminator and the generator enhances that from \texttt{stylegan2}~\cite{Stylegan2}, where the penalty of the discriminator enforces smoother separation of classes. In our approach, the GAN is \textit{conditional}, meaning that the discriminator takes two inputs (an image and landmarks), and by doing so, ensures that the fake image depends on the landmarks.   
In generator optimization, we sum the original generator loss ($\mathcal{L}_g$) with the loss between the restored and the initial images ($\mathcal{L}_{rec}^I$).
We train the GAN together with the style encoder (necessary for the style adjustments to assure that the style of the fake looks similar to both the generated style and to that of the transformed fake).

\begin{algorithm}[b]
\SetAlgoLined
 \textbf{Input}: image $I$\;
 encode coordinates of landmarks: $X^{l} = X^{l}(I)$\;
 compute heatmap of landmarks: $L = L(X^{l})$\;
 initialize $g$: random geometric transformation\;
 generate fake: $I_F = G(L, S_o(z))$\;
 restore image: $I_R = G(L, S(I))$\;
 landmarks of fake: $L_F = L(I_F)$\;
 $\min_{L} \left\{ 
 \begin{array}{cc}
    c_3 \mathcal{R}_{\flat}(X^{l}) + c_4 \mathcal{R}_g(g, I, L, X^{l}) + \\
    c_5 \mathcal{L}_g(I_F) + c_6 \mathcal{L}_{rec}^{I}(I_R, I) + c_7 \mathcal{L}_{rec}^{L}(L_F)
 \end{array}
 \right \}$\
 \caption{Single iteration of landmark encoder optimization}
 \label{encoder_train}
\end{algorithm}

 In Algorithm~\ref{encoder_train}, the resulting landmarks, produced by the encoder, have coordinates $X^l$ and the heatmap $L$, computed from $X^l$ by means of expression (\ref{hm_eq}). 
The encoder is optimised by the landmarks' participation in the generation of the fake and the restored images.
Regularizers $\mathcal{R}_{\flat}$ (distance to the barycenter), $\mathcal{R}_g$ ($g$-transform synchronisation), and $\mathcal{L}_g, \mathcal{L}^I_{rec}, \mathcal{L}^L_{rec}$ (influence of the decoder) are implemented exactly as described above in Section \ref{loss_func_sec}. To make training procedure more stable, we use accumulation of weights (moving average of weights over iterations). At each iteration, weights of the encoder model are multiplied by some decay coefficient and summed with the accumulating weights. After some number of iterations, the encoder replaces its weights by these accumulated values.


\section{Conditional GAN Architecture}
To build the conditional GAN for our purpose, we enhance the \texttt{stylegan2} architecture \cite{Stylegan2} by introducing the following modifications to the original generator and discriminator (see Supplementary Fig. 1 for details). 

\paragraph{Generator.} We consecutively downsample the heatmap of the landmarks by convolutional layers from the size $256\times256$ to the sizes $[4\times4, 8\times8, \ldots, 256\times256]$, and then, we concatenate them with the outputs from the progression of the modulated convolution blocks (\texttt{ModulatedConvBlock}) in network $G$. Term modulated convolution means that its weights are obtained from style. The noise $z$ passes through a series of linear layers to place the style $S_o(z)$ on the manifold. These styles are further used in \texttt{ModulatedConvBlock} to obtain the corresponding modulated weights. So, each \texttt{ModulatedConvBlock} receives pre-processed heatmap, multiplies it by the  modulated weights, does some non-linear transform, and passes it to the next block. The last block returns the fake image. 

\paragraph{Discriminator.} In the discriminator, we first apply convolution to the landmarks and image separately, concatenate them by channels dimension, and then pass it through a sequence of \texttt{ResNet} layers. 

\paragraph{Landmark Encoder.} Our Landmarks encoder consists of two principal parts.
Due to recent success of the stacked hourglass model~\cite{Hourglass}, we have integrated it in our landmarks encoder, which produces an output with separate channels corresponding to key-points. Consequently applying downsampling convolutions, we produce coordinates of the landmarks~($X^l$).

\paragraph{Style Encoder.} Style encoder is just a regular CNN that maps image to style matrix of size $2\times512$. The first matrix row is for low resolution \texttt{ModulatedConvBlock} layers, the second matrix row is for high resolution blocks, starting from $16 \times 16$.
\section{Experiments}
\label{experiments}%

\paragraph{Datasets.} We consider three popular face datasets: CelebA \cite{celebadataset}, 300-W \cite{w300dataset}, and MAFL \cite{MAFLDataset}. The CelebA dataset lacks the ground-truth landmarks, 300-W dataset contains 68 annotated key-points per face, having \num[group-separator={,}]{3000} training pairs in total. MAFL dataset has \num[group-separator={,}]{19000} training pairs (\num[group-separator={,}]{1000} test pairs); but, historically, it was annotated using only 5 key-points per face instead of 68. 

\begin{figure*}[h]
\begin{center}
\includegraphics[width=0.99\textwidth]{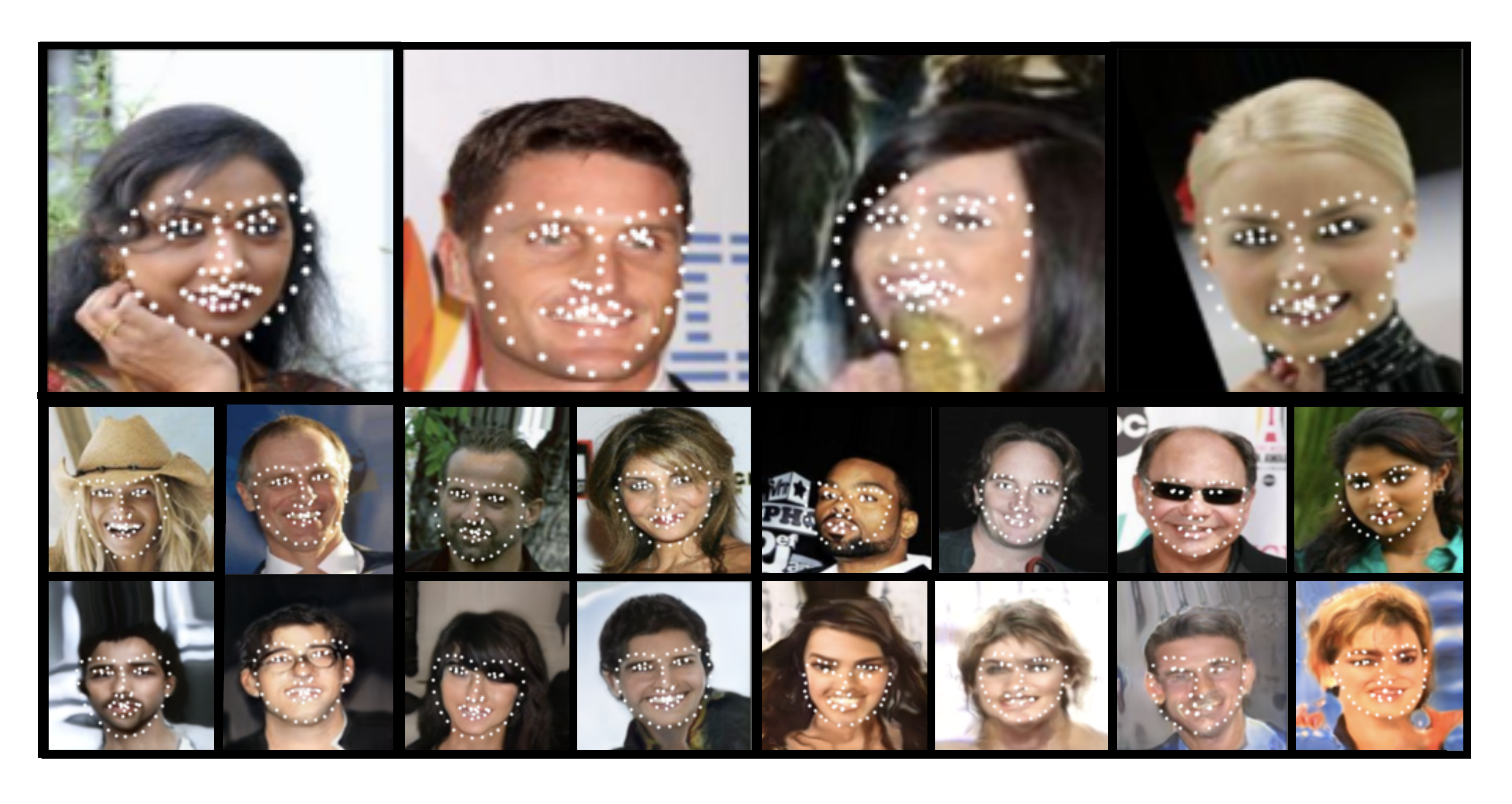}
\caption{Unsupervised BRUL\`E. The first and the second rows are real images with predicted landmarks \textit{without any supervision}, the third row is generated (conditionally) by GAN from the same landmarks as in the photos directly above.}
 \label{fig:true_fakes}
\end{center}
\end{figure*}

These three datasets are quite similar; however, the difference in the number of key-points, in the resolution, and in the zooming scale of the face images, provides a good setting for testing robustness and universality of our method.
As described above, we aspire to accomplish a completely \emph{unsupervised} extraction of landmarks using the BRUL\`E framework; but we also demonstrate efficacy in the \emph{semi-supervised} training scenario so that we can compare to the state-of-the-art methods. We compare supervised, semi-supervised, and unsupervised approaches in Table \ref{yaha_tab}.
\begin{table}[b]
  \begin{tabular}{ | p{60pt} | r | c | c |}
\hline
\textsc{Method} & \textsc{Training} & \textsc{MAFL} & \textsc{300-W} \\ \hline
TCDCN \cite{TCDCN} & \textit{supervised} & 7.95 & 5.54 \\ 
RAR \cite{RAR} & \textit{supervised} & - & 4.94\\
WingLoss \cite{WingLoss} & \textit{supervised} & - & 4.04\\
HG2 \cite{Hourglass} & \textit{supervised} & 2.3 & 4.2 \\
UDIT \cite{UDIT} & \textit{unaligned} & - & 5.37\\
Our method & \textit{semi-sup}. & \textbf{2.1} $\pm$ 0.03 & \textbf{3.9} $\pm$ 0.07 \\ \hline
Sparse \cite{SPARSE} & \textit{pre-train}  & 6.67 & 7.97 \\
Str. Repr. \cite{structrepr} & \textit{pre-train} & 3.15 & - \\
Fab-Net \cite{FabNet} & \textit{pre-train} & 3.44 & 5.71\\
Dense 3D \cite{Dense3D} & \textit{pre-train} & 4.02 & 8.23\\
DVE HG \cite{DVE} & \textit{pre-train} & 2.86 & 4.65\\ \hline
Our method & \textit{unsup.} & \textbf{8.8} $\pm$ 0.2 & \textbf{12.4} $\pm$ 0.2\\ \hline 
\end{tabular}
\caption{Quantitative comparison on MAFL and 300-W datasets using IOD (\%) as a metric. Semi-supervised (\textit{semi-sup.}) means a method is trained on MAFL or 300-W and uses additionally unlabelled images from CelebA dataset. \textit{Unaligned} denotes training cycle GAN with unaligned dataset. Unsupervised pre-trained (\textit{pre-train}) methods use full dataset to pre-train the encoder. To the contrary, our unsupervised (\textit{unsup.}) method needs just a single barycenter which could be computed on 10 face images.}%
\label{yaha_tab}
\end{table}

\paragraph{Training Scenarios: Unsupervised \textit{vs.} Pre-training \textit{vs.} Semi-supervised.} We follow the terminology from \cite{Qi-Unsup-Semisup-Review} and \cite{Chapelle-semisup-review}, where \textit{pre-trained} encoders are deemed as models that are not truly unsupervised. The reason is that such models (papers \cite{SPARSE,structrepr,FabNet,Dense3D,DVE,UDIT} for faces, in particular) use rather large datasets to pre-train their encoders (\num[group-separator={,}]{3000} images on W300 and \num[group-separator={,}]{19000} images on MAFL). Their performance on unseen data is then impressive, but can hardly be called unsupervised when so many images were needed to pre-train.
To the contrary, our approach requires only $10$-$20$ landmark images merely to calculate their barycenter, which is separate from the training routine and is unaffected by the size of the full dataset. 

Below, we report our experiments in \textit{unsupervised}\footnote{A single barycenter computation is needed.} and \textit{semi-supervised}\footnote{The annotation from the MAFL/300-W training sets are used, but the loss is optimized only on unlabeled dataset CelebA.} scenarios on exactly the
same data as the other works. There is no other unsupervised key-points extraction methods in the literature, according to definitions in \cite{Chapelle-semisup-review, Qi-Unsup-Semisup-Review}, and thus, there is no fair comparison for this case. 
Our semi-supervised BRUL\`E statistically outperforms all pre-trained and even some supervised methods (Table \ref{yaha_tab}).
Given that semi-supervised BRUL\`E works better than all pre-trained models, we do not consider the pre-training of our model\footnote{The bottleneck of BRUL\`E's encoder distills the landmarks by its conceptual design; so, \textit{pre-training} it would imply training the entire BRUL\`E.}.

\subsection{Unsupervised Experiments} In the unsupervised scenario, we first compute the  barycenter using landmarks from 50 images in the 300-W (or MAFL) datasets. 
We stress that this initiation does not contradict the criteria for being unsupervised \cite{Qi-Unsup-Semisup-Review}, because by doing so we essentially only `show' the object of interest to the model.
Moreover, our experiments have verified that one needs only $10$-$20$ landmark annotations to compute an accurate barycenter (refer to Fig.~\ref{fig:sample_size} below). 
Once computed, the barycenter is kept fixed for all experiments. 

The training follows the recipe from Section~\ref{method_sec}, with the Algorithm~\ref{gan_train} hyperparameters being set to $c_1 = 30$, $c_2 = 10$ and $\lambda = 80$. 
Hyperparameters\footnote{Found with greedy search algorithm and patience. We make no claims about a local minimum set by these (functional) hyperparameters.} in Algorithm~\ref{encoder_train} are initiated as: $c_3 = 14$, $c_4 = 1500$, $c_5 = 0.4$, $c_6 = 40$ and $c_7 = 20$. Parameters in $\mathcal{R}_\flat$ are: $c_{1}^{\flat} = 1$, $c_{2}^{\flat} = 2$, and $c_{3}^{\flat} = 4$. Parameters in $\mathcal{R}_g$ are: $c_{1}^{g} = 0.001$.
The error values of the landmark prediction were evaluated on the standard test sets of 300-W and MAFL, using a conventional metric -- the inter ocular distance (IOD)~\cite{interocular_distance}.
However, during the training, instead of the Euclidean norm like in the other works, we used the Wasserstein $W_1$ distance~\cite{bc_coords_proj} between the predicted and the ground truth landmarks (divided by the distance between the key-points corresponding to the outer eye corners), because the key-points lack the pairwise correspondence. The chosen metric for the test set (IOD) was identical for all methods in our comparison.

Performance of our method is demonstrated in Fig.~\ref{fig:true_fakes} and the comparison against the state-of-the-art methods is summarized in Table~\ref{yaha_tab}. Training of the model takes three days on three V100 Tesla GPUs, which is of the same order of magnitude as the other GAN-based models and the ones in the Table~\ref{yaha_tab}.
 
\subsection{Semi-Supervised Experiments}
\begin{figure}[t]
\includegraphics[width=0.9\textwidth]{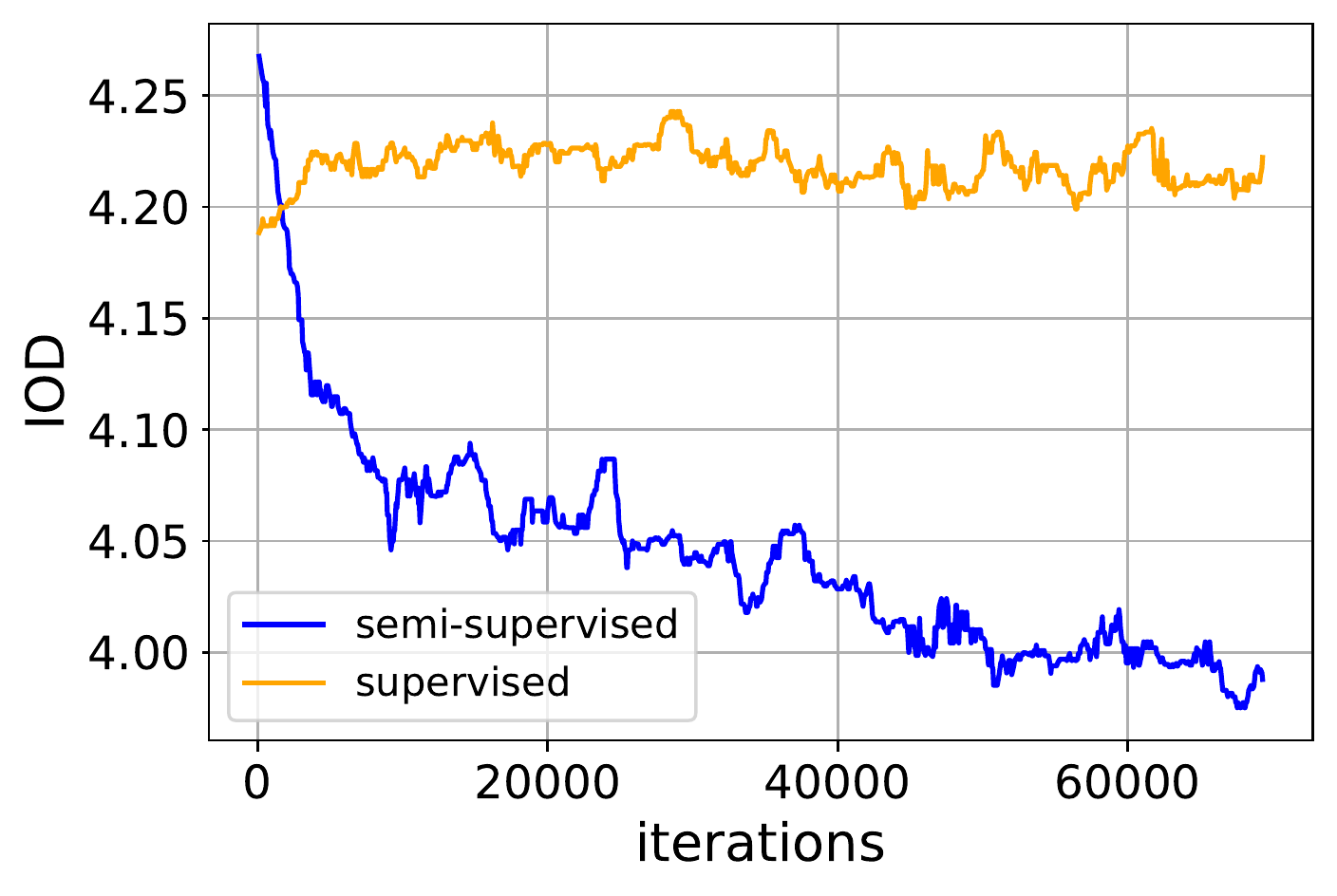}
\caption{Regularizer for supervised training on 300-W dataset. Orange line shows IOD metric for training HG2 \cite{Hourglass} model from a pre-trained state. Blue line corresponds to training \textit{the same model} with additional BRUL\`E loss fine-tuned on CelebA.}
\label{fig:hg_sup_train}
\end{figure}

In the semi-supervised case, we extend the unsupervised setting by using the training sets from the 300-W and the MAFL datasets. Thus, we make additional optimization using the landmarks from the annotation and the binary cross entropy loss. Note, however, that we optimize BRUL\`E losses only on the CelebA dataset, keeping the same hyperparameters for the semi-supervised experiments. As such, the BRUL\`E pipeline acts as a generic regularizer in the semi-supervised training, increasing stability and reducing the overfitting, which could be clearly seen in the training curve shown in Fig.~\ref{fig:hg_sup_train}. Visual quality of the semi-supervised BRUL\`E results can be assessed in Fig.~\ref{fig:supervised_one_style}, where we also show the content and style decomposition.

\subsection{Ablation Study} 
We have tested all of our loss terms to evaluate their individual contribution to the prediction accuracy. The results are shown in Fig.~\ref{fig:brule_h_losses}. We sequentially turned off the regularizers $\mathcal{R}_{\flat}$ and $\mathcal{R}_{g}$. The baseline value is the the black line which is the landmarks predicted by the mean value (the barycenter). BRUL\`E with omitted regularization $\mathcal{R}_{\flat}$ (pink curve) fails to reach the region close to the barycenter. BRUL\`E with omitted $\mathcal{R}_{g}$ (green curve) is already capable of improving the IOD beyond the barycenter, but does not control the scale of the deviations. The observed dynamics confirms our initial suggestion that these two regularizers perform best when they are combined. It is important to note that the ablation experiments were conducted with enabled GAN losses, because our GAN plays the role of decoder, without which our architecture would lose the image-landmarks correlation.

\begin{figure}[h]
\includegraphics[width=0.9\textwidth]{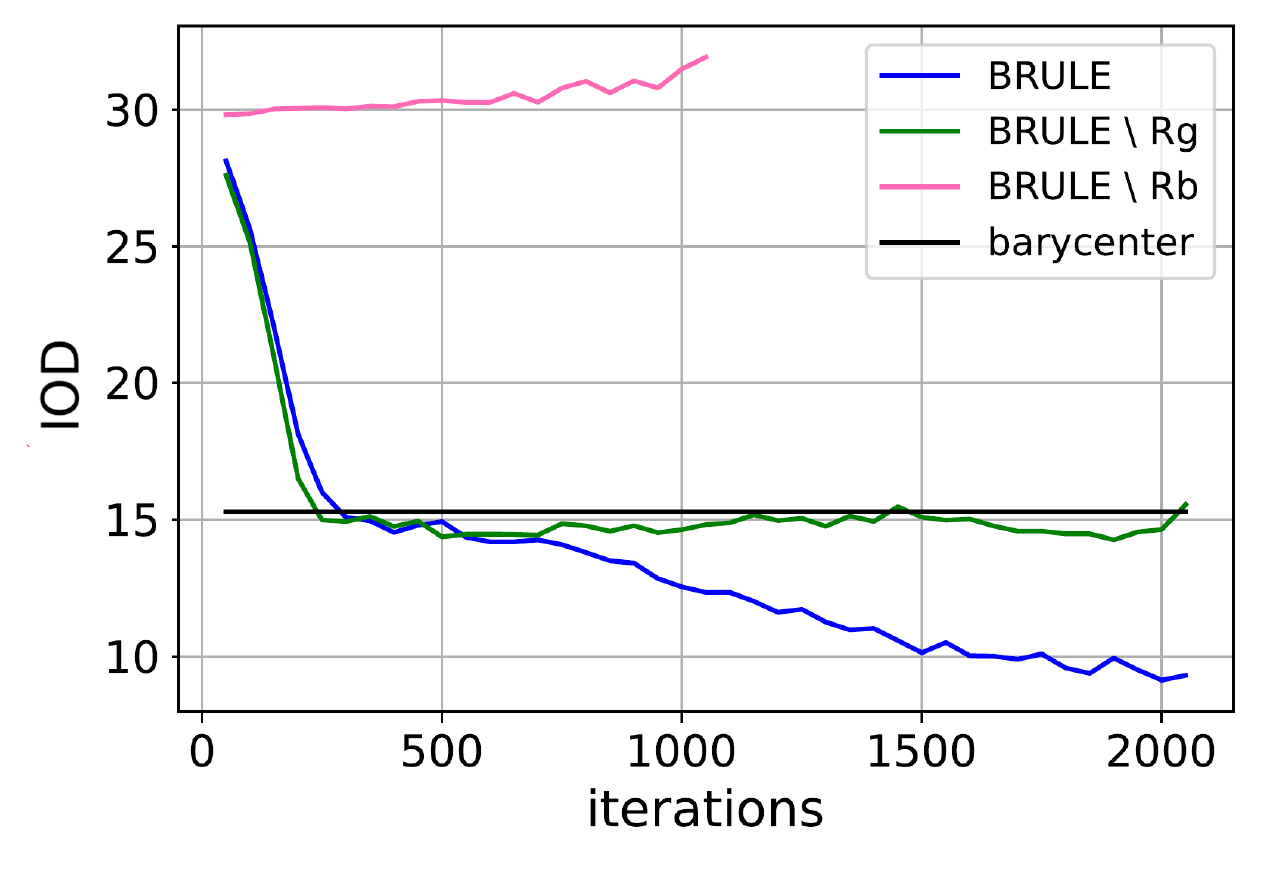}
\vskip -5pt
\caption{
BRUL\`E ablation study. Notice that when geometric transform regularizer $\mathcal{R}_{g}$ is switched off, the curve is at the level of the barycenter; whereas, switching off the barycenter regularizer $\mathcal{R}_{\flat}$ may even lead to divergence. Combined together, these two regularizers allow to improve beyond the barycenter (blue curve).
}
\label{fig:brule_h_losses}
\end{figure}

\begin{figure}[h]
\includegraphics[width=0.95\textwidth]{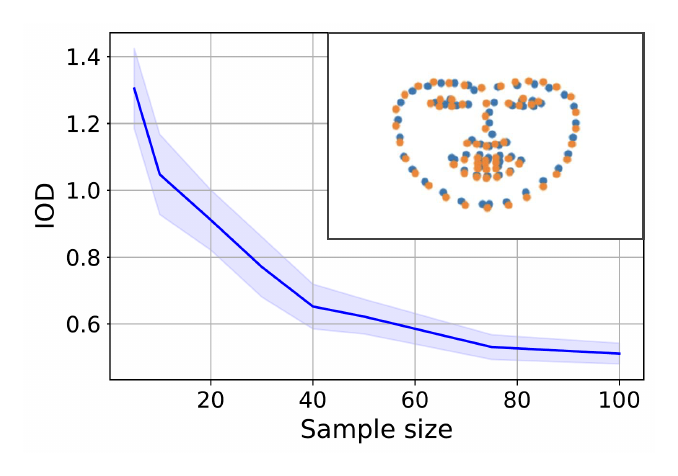}
\caption{IOD error as a function of number of faces used to compute the barycenter. The inset visually compares the reference barycenter (all faces) to the empirical one ($10$ faces). }
\label{fig:sample_size}
\end{figure}

\begin{figure*}[t]
\begin{center}
\includegraphics[width=0.95\textwidth]{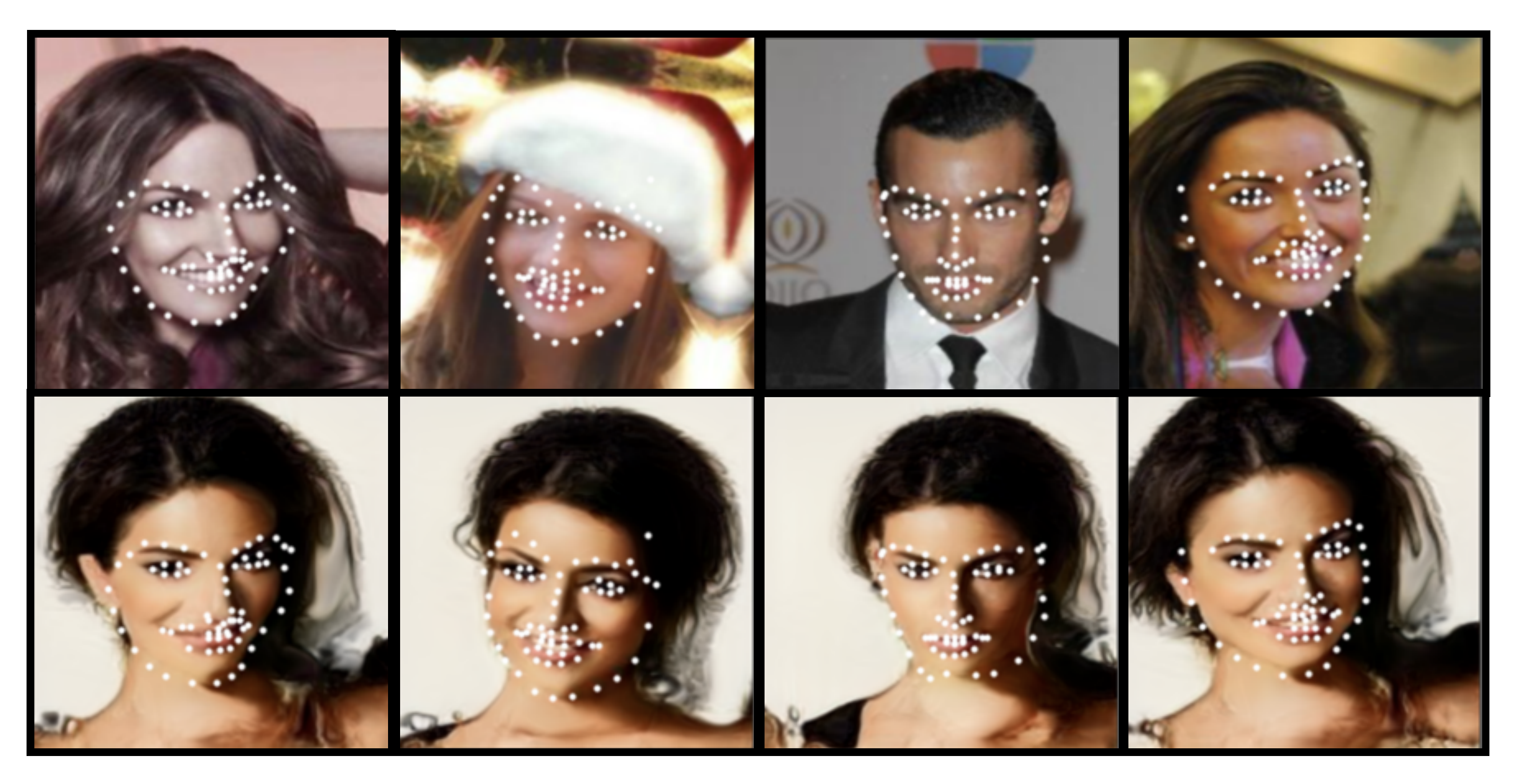}
\caption{Semi-supervised BRUL\`E. \emph{Top row}: real images with predicted landmarks. Our regularization improves the detection both quantitatively (see Table~\ref{yaha_tab}) and perceptually.  \emph{Bottom row}: fake images generated by conditional GAN with the same landmarks as in images above and a new style.
 \label{fig:supervised_one_style}}
\end{center}
\end{figure*}

Finally, we have also studied the influence of including various numbers of landmark images into the calculation of the barycenter on the performance metric (IOD). Fig. \ref{fig:sample_size} demonstrates the IOD distance to the reference barycenter depending on the sample size. The inset in the figure portrays two barycenters: the first one is calculated using just 10 samples, and the other one is a result of the calculation using the entire dataset (the reference barycenter). This figure shows a hyperbolic attenuation of the IOD error. Note that even 10 annotated landmarks generate a reliable barycenter, with the error being rather small in the absolute values. Even when reduced to the critical level of only 5 samples, the resulting error and stability are observed to be acceptable.

\section{Discussion}
Face landmarks predicted by our unsupervised method, as illustrated in Fig.~\ref{fig:true_fakes}, are very good outcomes, especially in the examples with the faces photographed from the front. 
Because this is the predominant orientation of the face in both datasets, the barycenter really 'looks' like the frontal photograph. Some problems emerge when the testing images are acquired from the side-view, probably requiring larger warp deformations or more complex affine transformations to compensate for the `sub-optimal' orientation.
However, the encoder captures all the landmark transformations \emph{only} in the 2D image plane, because the geometric transforms used in $\mathcal{R}_g$ and in the discriminator are both 2D transforms. Hence, the future work will entail extension of these transforms into 3D domain, which is expected to significantly boost performance on those faces that look up/down/sideways or have a weird viewangle.

Moreover, Fig. \ref{fig:supervised_one_style} shows first adaptation of \texttt{stylegan2} to conditional generation. In the semi-supervised case, it splits the landmarks and the style data exceptionally well, so that we can generate fake photos of a person (the fixed style) from different sets of landmarks. We find BRUL\`E to be well positioned for regularization in active learning (AL) frameworks to gain efficient annotation strategies~\cite{Wang-AL,shelmanov2019active}. AL and multi-class / manifold barycenters \cite{bc_gar} are both obvious extensions for the future work.

\section{Conclusions and Broader Impact}
We demonstrate the efficiency of our unsupervised segmentation method on the datasets that contain images of faces, with the segmentation implying extraction of the facial landmarks. One of our `firsts' is that this key-point extraction is \emph{100\% interpretable}\footnote{BRUL\`E's bottleneck contains nothing but the landmarks by its design.}. An immediate practical impact is expected in the areas of video and image editing, where detection/selection of the landmarks could be done automatically. Painting over an object, changing the style of an object in a scene, combining various objects from different sources in one picture -- all of these areas have acquired a powerful tool into their arsenal. The proposed method will be useful in all tasks where a large amount of unlabelled data prevails the labeled samples, and also in semi-supervised tasks where one wants to reduce overfitting by the unlabelled dataset. 

As an anticipated impact of our work, BRUL\`E will become very useful in the biomedical domain. In the clinical setting, where one requires high accuracy of the per-pixel image predictions to support the diagnostic decisions, only specialists from the field can perform the essential annotations, significantly slowing down the time for obtaining a marked dataset. However, our method relies on the use of barycenters (the `average image' values), effectively enabling the solution: because the same organ is quite similar among different patients, computing the average barycenter is a very sensible endeavor to generalize among large patient cohorts. Our method will change the way the segmentation is approached when there is no or limited annotation data, paving the way towards the ultimate vision of complete image understanding with no supervision.
%
\section{Acknowledgements}
We thank Ivan Oseledets and Victor Lempitsky for helpful discussions.

{\small
\bibliographystyle{main}
\bibliography{main}
}
\begin{figure*}[b]
\begin{center}
\includegraphics[width=\textwidth]{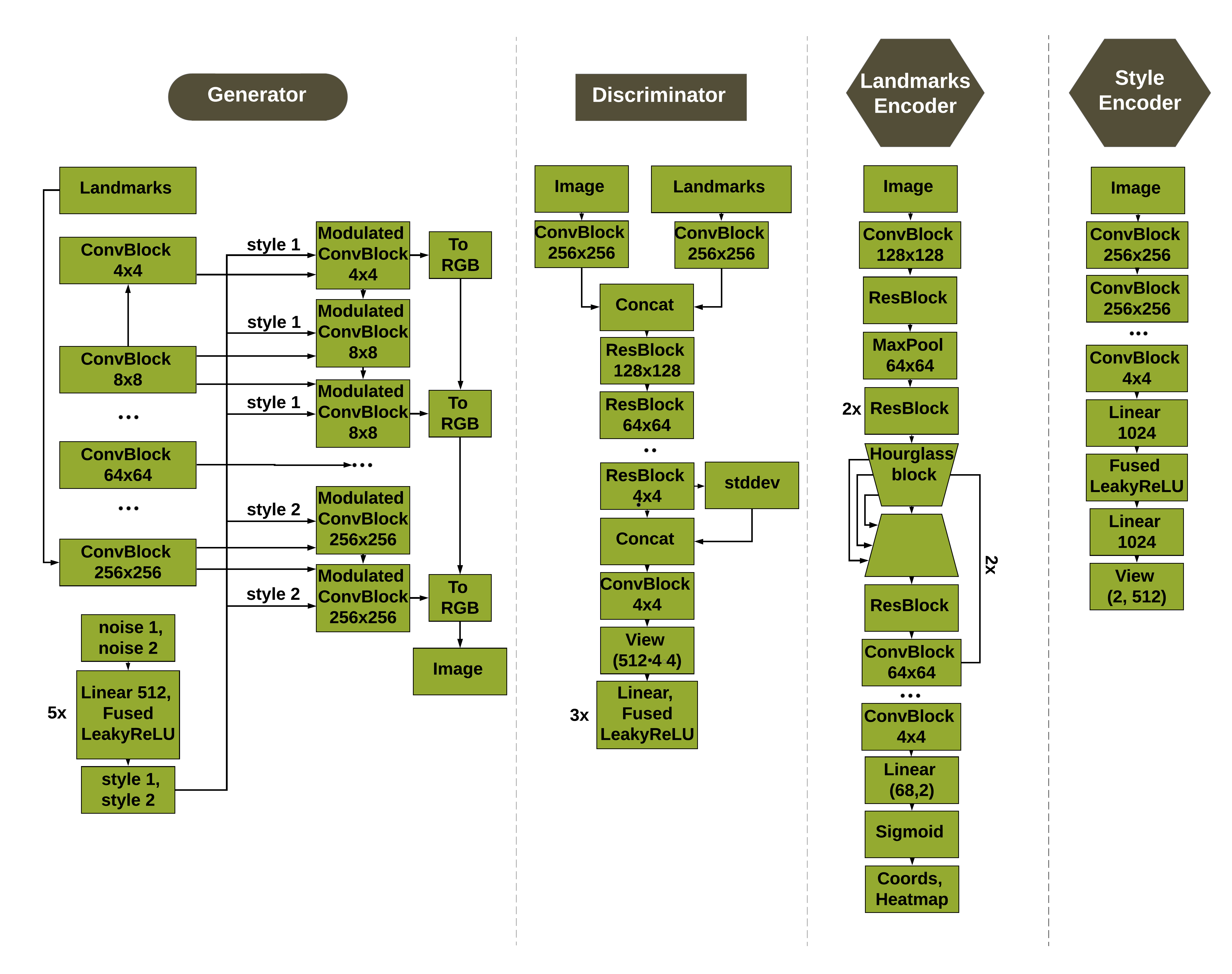}
\caption{Complete architecture. \texttt{ConvBlock} is a combination of convolution layer and \texttt{LeakyReLU}. Blocks \texttt{Fused LeakyReLU}, \texttt{ModulatedConvBlock}, \texttt{To RGB} are borrowed from \texttt{stylegan2}.
}
\label{fig:cont_style_encoders}
\end{center}
\end{figure*}

\begin{figure*}[hbt]
\begin{center}
\includegraphics[width=\textwidth]{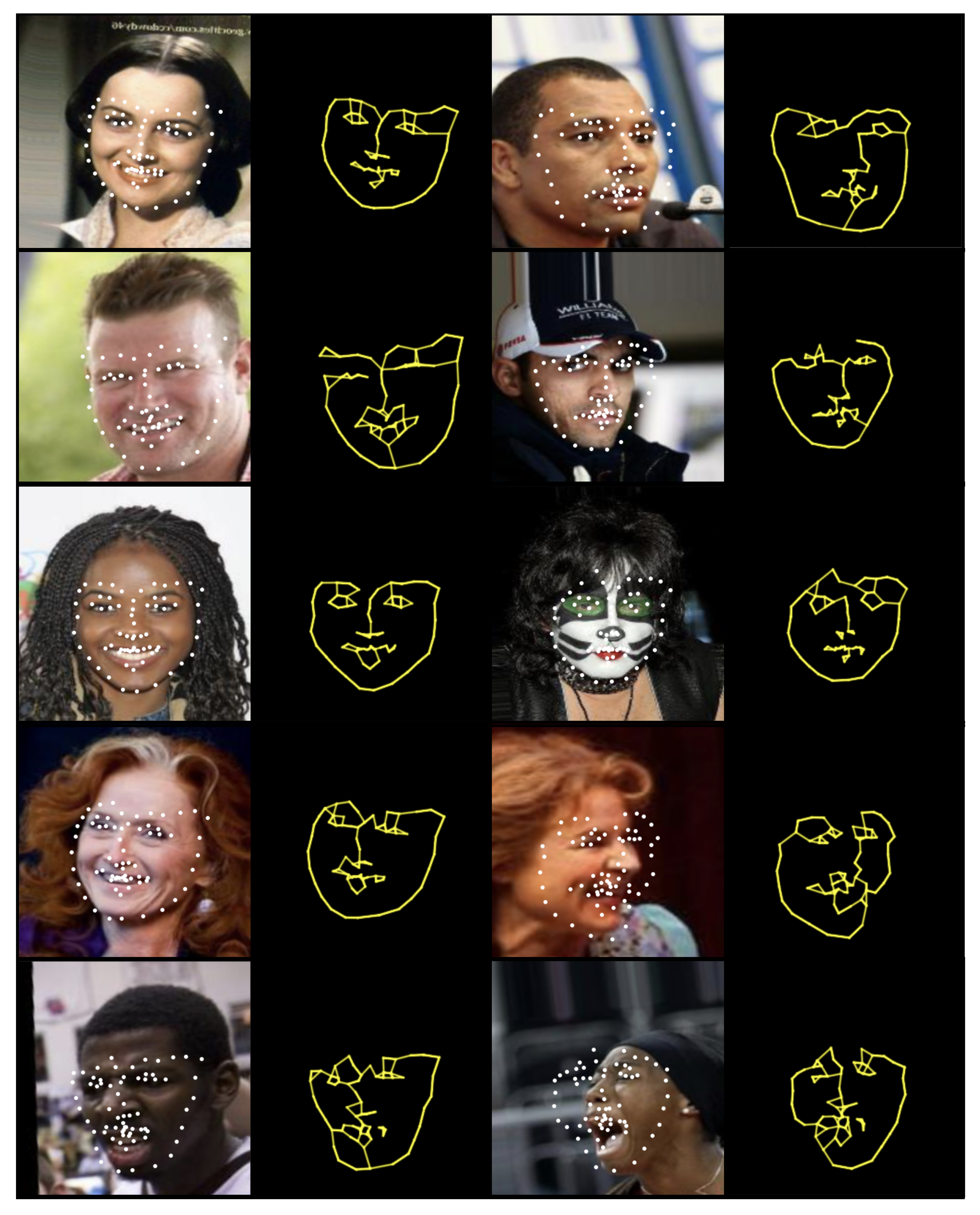}
\caption{Unsupervised BRUL\`E. \textit{Columns 1 and 3}: images from CelebA dataset with detected landmarks. \textit{Columns 2 and 4}: corresponding skeleton heatmaps described in the main text.
}
\label{fig:radost_perfectzionista}
\end{center}
\end{figure*}

\begin{figure*}[hbt]
\begin{center}
\includegraphics[width=\textwidth]{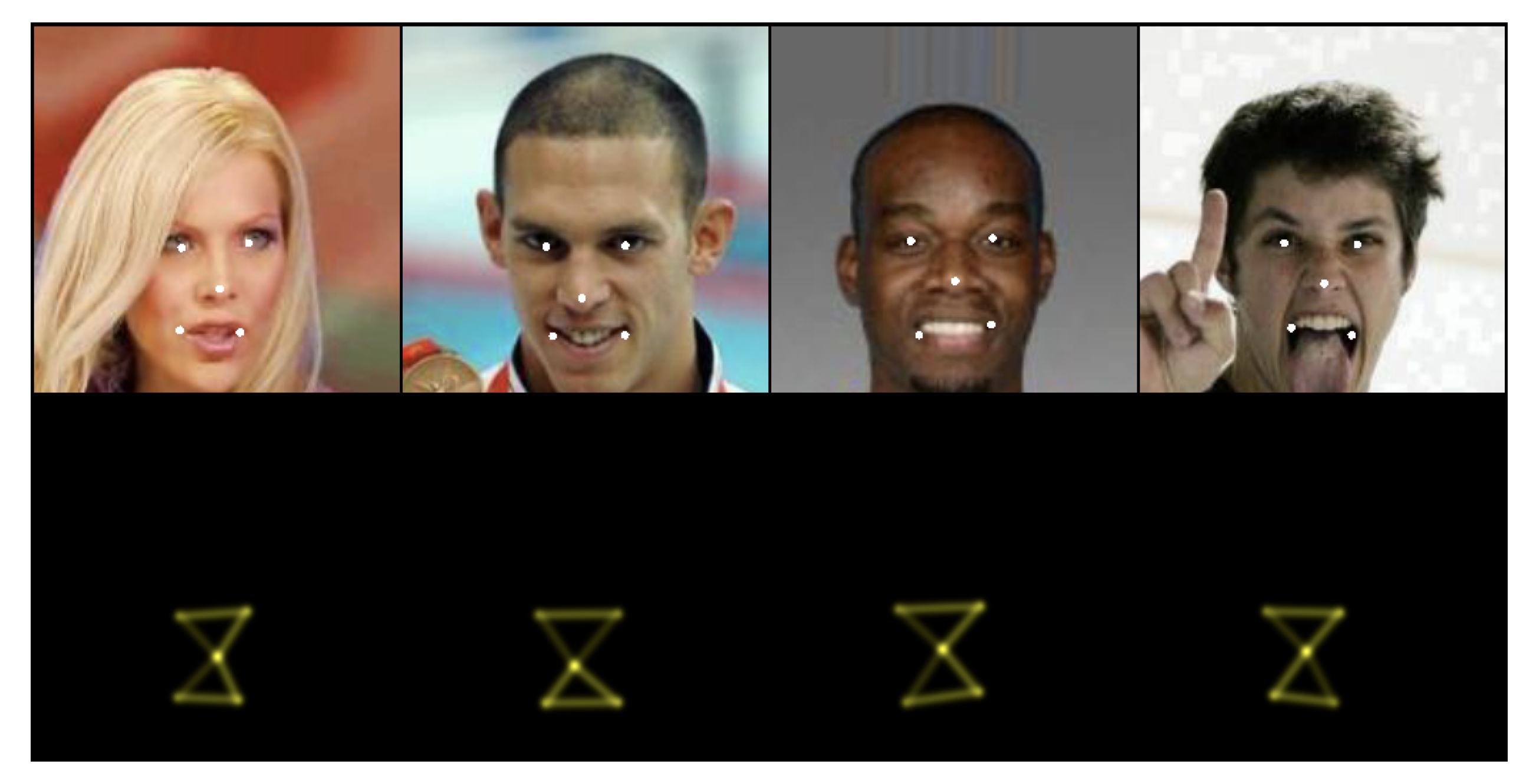}
\caption{Unsupervised BRUL\`E. \textit{First row}: images from MAFL dataset with detected landmarks. \textit{Second row}: corresponding skeleton heatmaps.
}
\label{fig:sandclock}
\end{center}
\end{figure*}

\begin{figure*}[hbt]
\begin{center}
\includegraphics[width=\textwidth]{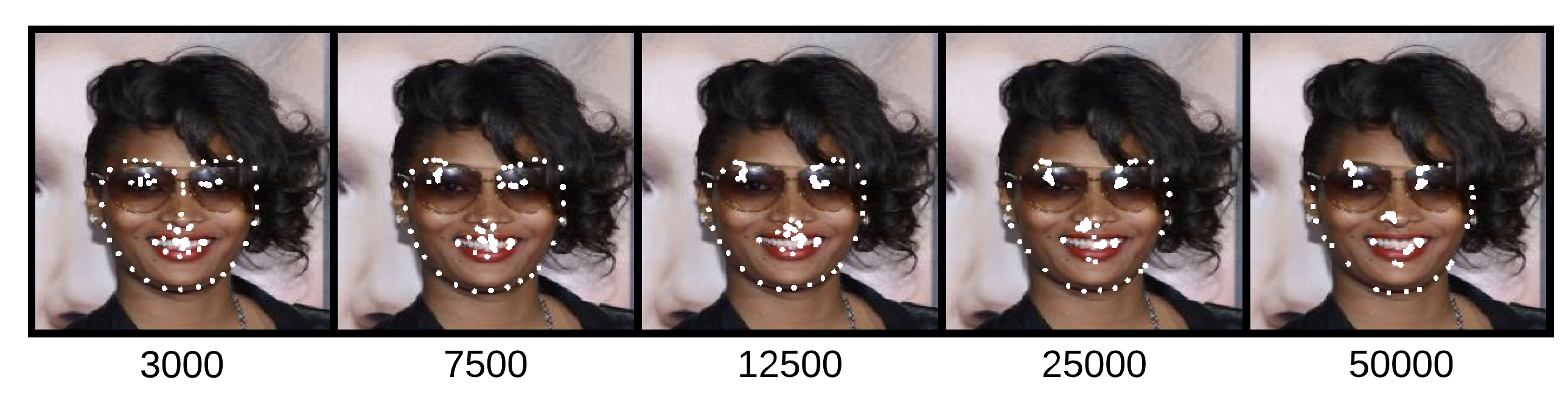}
\caption{Example of the collapse of landmarks with increasing strength of $\mathcal{R}_g$.  The frame labels correspond to the coefficient by $\mathcal{R}_g$. This component eliminates the collapse of the landmarks into the barycenter, but it may produce clusters of points, as shown in this example.
 \label{fig:g_collapse}}
\end{center}
\end{figure*}

\begin{figure*}[hbt]
\begin{center}
\includegraphics[width=\textwidth]{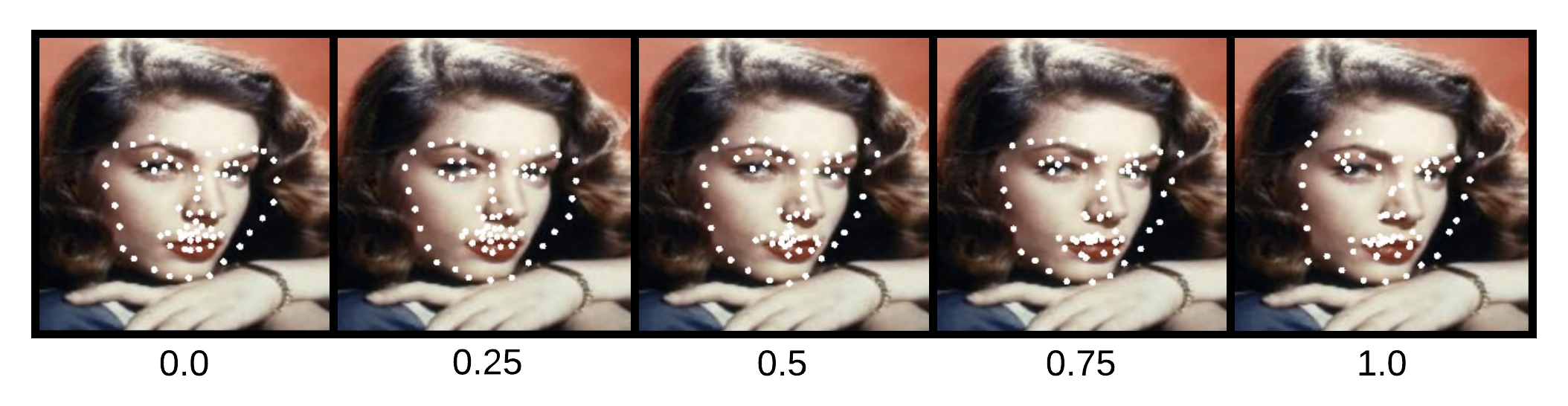}
\caption{Example of the influence of the generator loss ($\mathcal{L}_g$). If the value of the coefficient is small, the facial details are not reflected well. The large values, on the other hand, destruct the landmark structure.
 \label{fig:olya_lena_glasha}}
\end{center}
\end{figure*}

\begin{figure*}[hbt]
\begin{center}
\includegraphics[width=\textwidth]{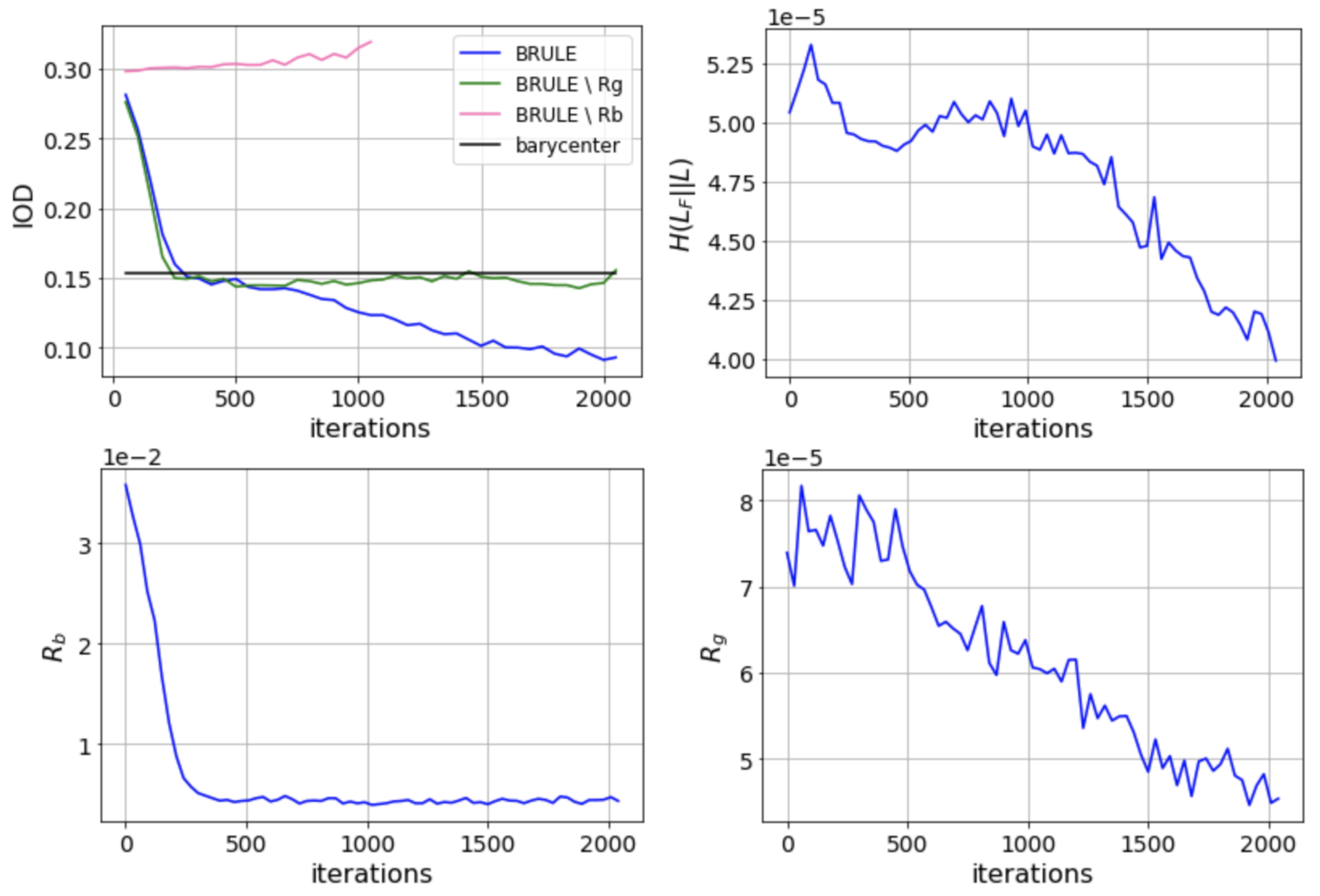}
\caption{
BRUL\`E ablation study using Inter Ocular Distance as the metric (IOD).
\newline\textit{Top left:} The value of the barycenter is the black curve. BRUL\`E with omitted regularization $\mathcal{R}_{\flat}$ fails to reach the barycenter. BRUL\`E with omitted $\mathcal{R}_g$ is already capable of improving the IOD beyond the barycenter (green). All six components of the loss function contribute to the landmark detection performance beyond the barycenter (blue). 
\newline\textit{Top right:} Behaviour of individual loss contribution of cross-entropy $H(L_F\|L)$ between the fake and the predicted landmarks.
\newline\textit{Bottom row:} Individual contributions of the regularizing loss components $\mathcal{R}_{\flat}$ and $\mathcal{R}_g$. 
 }
\label{fig:brule_h_losses}
\end{center}
\end{figure*}


\end{document}